\documentclass[a4paper,twocolumn,fleqn]{article} 

\usepackage{ist}
\usepackage[super]{nth}

\title{Revisiting and Optimising a CNN Colour Constancy Method for Multi-Illuminant Estimation}

\author{ Ghalia Hemrit$^{12}$ and Joseph Meehan$^1$ \\ $^1$Huawei Technologies, Nice, France; $^2$University of East Anglia, Norwich, United Kingdom}
\date{} 

\begin{document} 

\maketitle 

\thispagestyle{empty} 


\begin{abstract}

The aim of colour constancy is to discount the effect of the scene illumination from the image colours and restore the colours of the objects as captured under a `white' illuminant. For the majority of colour constancy methods, the first step is to estimate the scene illuminant colour. Generally, it is assumed that the illumination is uniform in the scene. However, real world scenes have multiple illuminants, like sunlight and spot lights all together in one scene. We present in this paper a simple yet very effective framework using a deep CNN-based method to estimate and use multiple illuminants for colour constancy. Our approach works well in both the multi and single illuminant cases. The output of the CNN method is a region-wise estimate map of the scene which is smoothed and divided out from the image to perform colour constancy. The method that we propose outperforms other recent and state of the art methods and has promising visual results.
\end{abstract}

\section{Introduction}
\label{sec:intro}


Performing colour constancy is an important pre-processing step of the digital camera pipeline. It consists of removing the colour bias introduced by the scene illuminant from the colours in the image. This way, a digital camera gives images of the scene where the objects have the same colours independently of the scene illumination conditions. Colour constancy is important for various computer vision tasks like object detection and tracking but can also be important to improve the visual aesthetics of an image. 

Most colour constancy methods rely on a first step which is estimating the colour of the scene illuminant before this colour can be discounted from the image. Very often, the single illuminant case is assumed and only the predominant light colour in the scene is estimated (global estimate) expressed as an RGB vector and used for colour constancy. This is the case for state of the art and most commonly used methods like statistics-based methods \cite{grayworld}\cite{whitepatch}\cite{cheng2014} and more recent learning-based methods \cite{CNNbianco}\cite{FC4}.     

This approach succeeds in solving for the illuminant colour in many cases and it has been largely adopted. There exist a wide variety of methods but the answer to illuminant estimation varies considerably with the method that is used \cite{survey} and can give unsatisfactory results, in particular when there are multiple illuminants in the scene. 

In fact, most real life scenes have more than one ambient light. We consider in this work the multi-illuminant case and propose instead of having a single (global) estimate for the scene illuminant colour, to generate multiple local estimates in the form of a region-wise map. This estimate map can be then divided out from the image to perform colour constancy. We use the fully convolutional neural network SqueezeNet-FC4 introduced by Hu et al. \cite{FC4}. Eponymously, we call our method the Multi-Estimate Map CNN (MEM CNN). We evaluate our method on the MIMO public multi-illuminant dataset for colour constancy \cite{Beigpour2014} and we compare it to other methods using the angular error metric that we adapt to the multi-estimate case. We show that MEM CNN is effective in the multi and single illuminant cases and that it outperforms the other methods when evaluated on this dataset of images.

The rest of the paper is organised as follows. In the next section, we present some related work on colour constancy and deep learning methods. Then, we describe the proposed approach and introduce the dataset that we used for training. We also present the network architecture in another section. This is followed by the experiments and results section, and a conclusion.
  
\section{Related Work}
Some recent learning-based methods address the question of the multiple illuminants in the scene and how to give an estimate of the illuminant colour in this particular case \cite{gijsenij}. Often, multiple local estimates are produced then combined into a global estimated illuminant. Additionally, when local estimates are used for colour constancy, the global estimate remain a reference in the colour difference evaluation.

In \cite{FC4}, local estimates are produced by the FC4 trained model as well as confidence values where a confidence expresses the value of the corresponding patch for estimating the illuminant based on its semantic content. The confidence-weighted local estimates are combined into a one global estimate.

In \cite{gijsenij}, the input image is sampled into patches and local estimates are produced with an edge-based method \cite{grayedge} then a clustering technique gives the global estimate. In \cite{cheng2016}, a user preference study allows to choose the global estimate. Here, estimates are predicted with a learning-based method from $4\times8$ sub-images and used to classify the image as single or two-illuminant. Then, two illuminants are calculated with a k-means clustering.

\cite{MIbianco} is a learning patch-based method. It also produces multiple local estimates. A detector decides whether the image has a single or multiple illuminants using a statistical analysis of the local estimates (the difference between the estimates in the chromaticity space is compared to a threshold). If the difference is small, it is a single illuminant case and a local-to-global regression is performed with a supervised neural network. In the multi-illuminant case, the local estimates are used with no further processing. The method was not evaluated in terms of colour differences in the multi-illuminant case.  In \cite{arxiv}, the local illuminant vectors are estimated from the whole image and not from patches. Instead of learning the estimates, the model predicts local kernels. These kernels are then used to calculate the estimated illuminant. A global-to-local aggregation gives a region-wise estimate map using clustering results. The evaluation in terms of colour differences uses the global estimate. 

We propose to use the estimate map from SqueezeNet-FC4 \cite{FC4} to perform colour constancy. The FC4 trained model generates 60 patch-based local illuminants organised in a $6\times10$ RGB estimate map. We upsample this map and process it to smooth the transitions between the colours. The map is then divided out from the image to restore the objects colours. Other methods also assume a smooth transition of the illumination in the scene image in the multi-illuminant case like \cite{retinex}\cite{ebner}. 


\section{Method Overview}

We present in this paper a framework based on a learning-based method for performing colour constancy in the multi and single illuminant cases. Our method, MEM CNN, uses the fully convolutional network SqueezeNet-FC4 introduced by Hu et al. \cite{FC4}. The original FC4 method gives a global estimate which is a single 3-colour vector. 

The novelty of our work is that we do not estimate a single illuminant in the scene. Instead, we use multiple estimates to solve for colour constancy. Our method generates an estimate map which is an image where the illuminant colour is estimated for every region of the image. 
 
The model generates an estimate map which is a thumbnail image of size $6\times10\times3$ (60 patches). By upsampling this image to the scene image size (by duplicating the patches), it becomes a region-wise estimate map. This map is linearised and more importantly, it is processed in a way that smoothes the edges between the different patches. To do that, we filter the map with a kernel of area $50\times50$. It is finally divided out from the image to remove the illuminant colour bias, as shown in Equation~\ref{eq:WB} for every pixel $k$ from the image. 

\begin{equation}
\centering
    \mathbf{I}_{c}(k)  = \mathbf{I}_{o}(k)diag(\hat{\boldsymbol{\rho}}(j))^{-1}
     \label{eq:WB}
\end{equation}

\noindent where $k$ refers to the $k$-th pixel from the scene image, in the $j$-th region. $\mathbf{I}_{o}$ is the image of the scene under illuminant $E_{o}$ and $\mathbf{I}_{c}$ is the restored image of the scene under white light $E_{c}$. $\hat{\boldsymbol{\rho}}(j)$ is the estimated illuminant RGB vector for the $j$-th region of the image, and $diag$ is the diagonal matrix operator.

\subsection{Training Dataset}

\begin{figure}
\centering    
\includegraphics[width=0.22\textwidth]{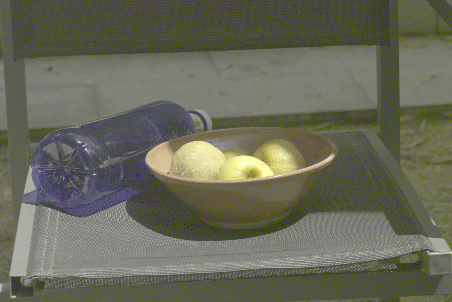}
\includegraphics[width=0.22\textwidth]{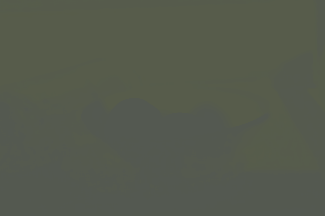}\\ 
\includegraphics[width=0.22\textwidth]{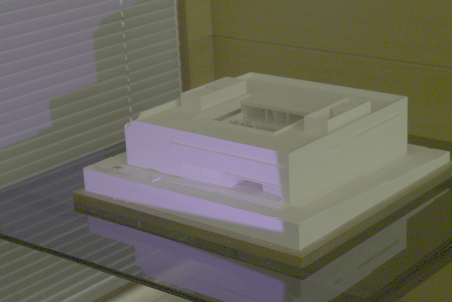}
\includegraphics[width=0.22\textwidth]{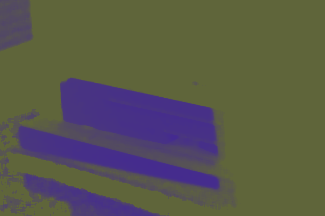}\\
\includegraphics[width=0.22\textwidth]{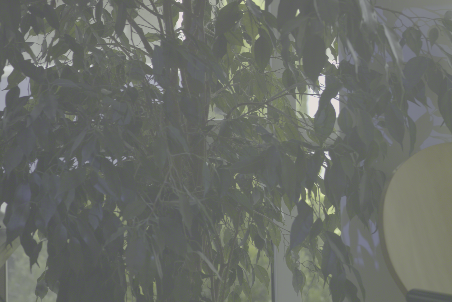}
\includegraphics[width=0.22\textwidth]{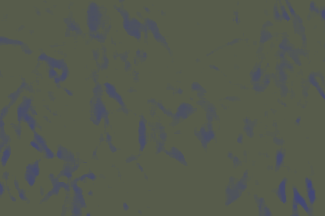}\\ 
\caption{Samples from the MIMO dataset \cite{Beigpour2014}: the captured scenes (gamma-corrected) and the corresponding pixel-wise ground-truth images.} 
\label{fig:dataset}  
\end{figure}

To train the network, we use the MIMO dataset introduced by Beigpour et al. \cite{Beigpour2014}\cite{MIMOsite}. This dataset has 78 RGB images of real life scenes and images taken in a controlled lab environment. The scenes have multiple illuminants: two main non-uniformly distributed illuminants in every scene. The dataset is different from other public sets for colour constancy as every image is provided with a pixel-wise ground-truth illuminants image which is also an RGB image that has the illuminant 3-vector for every pixel of the corresponding captured scene. Scenes images and pixel-wise ground-truth images have size $302\times452\times3$. Samples from the dataset are presented in Figure~\ref{fig:dataset}.

The scenes contain simple and complex diffuse and specular objects. The pixel-wise ground-truth is calculated following a methodology described in \cite{Beigpour2014} where the authors present a procedure to collect the data and calculate the ground-truth. This methodology is not in the scope of this paper.

The methodology is given for two illuminants scenes which is the case of the lab-based images. For these scenes, three coloured light sources were used: a reddish, a blueish and a `white' illuminants. Every scene was lit by a distinct pair of light sources from two angles: left and right. The real life scenes contain two major illuminants: the ambient light which is the overall illumination and a direct light which was added to the scene. For this reason, the given ground-truth calculation is also considered valid for these scenes and pixel-wise ground-truth for these scenes were calculated the same way. 

The neural network takes inputs of size $216\times 325\times3$. The dataset images are cropped (at random positions) to fit the input size requirement, as part of a data augmentation step. The images cannot be resized because the ground-truth image pixel values should not change. Data augmentation also includes random vertical and horizontal flipping of the training sample. The ground-truth image is processed accordingly. As the FC4 network was pre-trained on gamma-corrected data for display, we also apply a gamma to the scenes linear RGB images. 

\subsection{Network Architecture}

FC4 \cite{FC4} uses a SqueezeNet set of layers \cite{squeezenet}. The model generates two types of outputs: an estimate map and a corresponding confidence map which has the confidence weights of the patches in the illuminant estimation. The whole estimate map is used for training (to calculate the loss) and for testing. We do not use the confidence map in this work.

The network is implemented with Tensorflow and it is trained end-to-end by back-propagation. Adam optimiser is used with $learning~rate=3\times 10^{-4}$ and $batch~size = 16$. All the convolutional layers are fine-tuned from the pre-trained network (for image classification). The ReLu function is used for non-linear activation except for the final layer 7. 

\begin{figure*}[!ht]
\centering    
\includegraphics[width=1\textwidth]{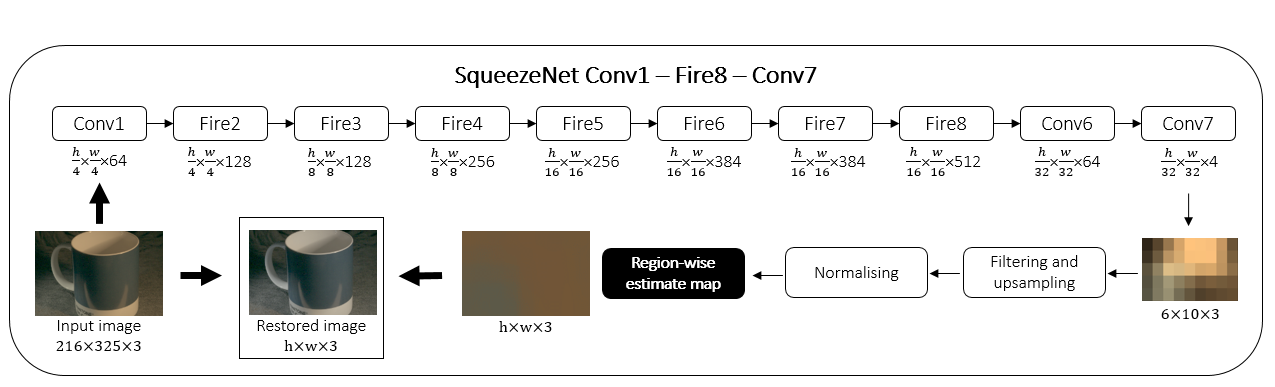}  
\caption{Architecture of MEM CNN.} 
\label{fig:architecture}
\end{figure*}

In MEM CNN we use the FC4 neural network. We present the architecture in Figure~\ref{fig:architecture}. The network is trained with the scenes images and ground-truth illuminants images. The input size is $216\times325\times3$.

\subsubsection{Angular Error Map}
We define a new cost function to take into account the multiple ground-truth illuminants in one image and the multi-estimate map. To calculate the loss for every training sample, the ground-truth illuminants image used for training is downsampled to the size of the estimate map ($6\times10\times3$): the image is divided into 60 distinct regions then the median over every region is calculated. For the $6\times10$ pairs of colour vectors, the traditional angular error is calculated. The loss $L$ is an average of the $60$ angular errors, as shown in Equation~\ref{eq:loss}. 

\begin{equation}
\centering
    L\left (\hat{\boldsymbol{\rho}}_{i} \right ) = mean(err(\hat{\boldsymbol{\rho}}_{i}(1), \boldsymbol{\rho}_{i}(1)), ..., err(\hat{\boldsymbol{\rho}}_{i}(60), \boldsymbol{\rho}_{i}(60)))
     \label{eq:loss}
\end{equation}

\noindent where $i$ refers to the $i$-th training sample, $\hat{\boldsymbol{\rho}}_{i}(j)$ is the $j$-th estimated illuminant which is the RGB vector of the $j$-th patch from the estimate map and $\boldsymbol{\rho}_{i}(j)$ is the $j$-th region-wise ground-truth illuminant. It is defined as the median  of the $j$-th region of the ground-truth image (see Equation~\ref{eq:mediangt}).

\begin{equation}
\centering
     \boldsymbol{\rho}(j) = median(\mathbf{R}_{j})
     \label{eq:mediangt}
\end{equation}

\noindent where $\mathbf{R}_{j}$ is the $j$-th region of the ground-truth image which is divided into $60$ equal regions of $36\times32$ pixels except the right side regions of the image which have more pixels.

We call the metric associated to the new cost function the angular error map as it uses the error distribution in a downsampled representation of the scene. The output of the angular error map is a single angular error measure. We use the angular error map when testing MEM CNN to evaluate its performance in terms of angular error. 

\section{Experiments and Results}
We evaluate the performance of our method MEM CNN and compare it to recent and state of the art methods on the MIMO dataset \cite{Beigpour2014}. 
We use the angular error as a metric. For MEM CNN, we first calculate the angular error maps. For the other single estimate methods, the estimate is an RGB vector, the angular error is the angle between the estimate and the ground-truth. For MIMO scenes images, we use a ground-truth RGB vector that we generate from the ground-truth pixel-wise image as being the average colour of the two RGB vectors with the largest weights in the image. This way, we define an illuminant colour vector per scene to use with the single estimate methods. Of course, the way this vector was defined from the image is questionable and there can be different calculation methodologies that can be adopted. Saying that, we tested other methods and we found very similar results. 

\subsection{Results}
 
\begin{figure}
\centering    
\includegraphics[width=0.33\columnwidth]{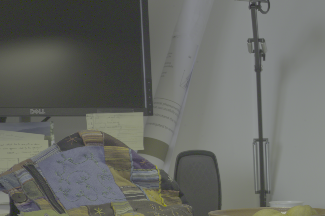}
\includegraphics[width=0.33\columnwidth]{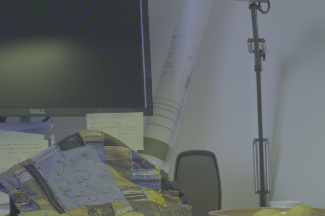}\\ \includegraphics[width=0.33\columnwidth]{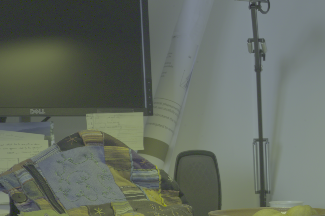}
\includegraphics[width=0.33\columnwidth]{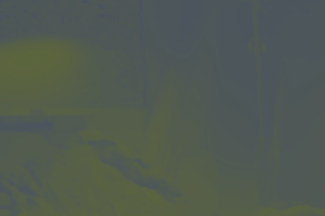}\\
\includegraphics[width=0.33\columnwidth]{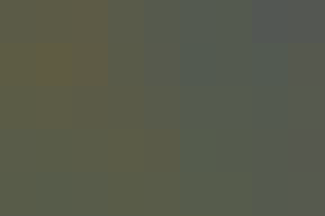}
\includegraphics[width=0.33\columnwidth]{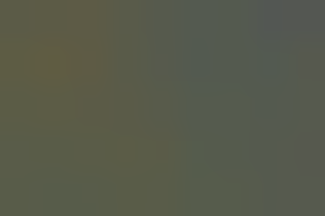}\\
\vspace{0.25cm}
\includegraphics[width=0.33\columnwidth]{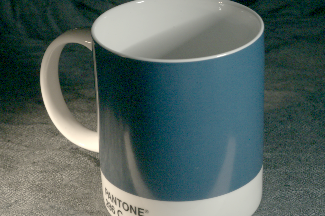}
\includegraphics[width=0.33\columnwidth]{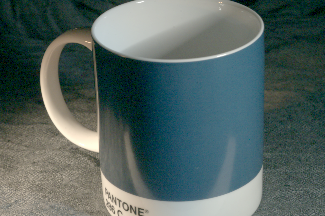}\\
\includegraphics[width=0.33\columnwidth]{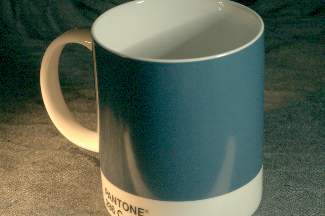}
\includegraphics[width=0.33\columnwidth]{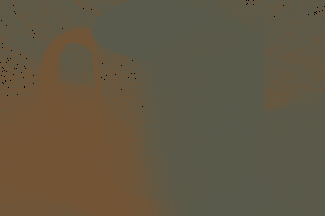}\\
\includegraphics[width=0.33\columnwidth]{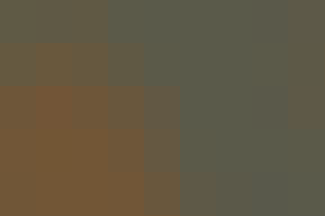}
\includegraphics[width=0.33\columnwidth]{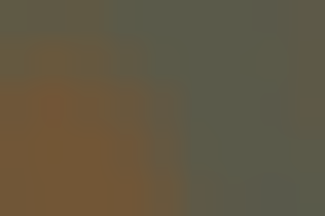}\\
\vspace{0.25cm}
\includegraphics[width=0.33\columnwidth]{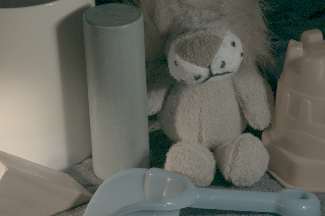}
\includegraphics[width=0.33\columnwidth]{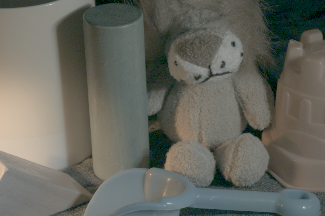}\\
\includegraphics[width=0.33\columnwidth]{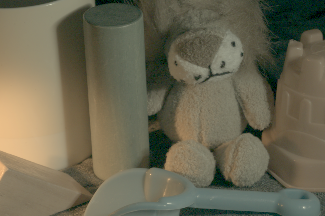}
\includegraphics[width=0.33\columnwidth]{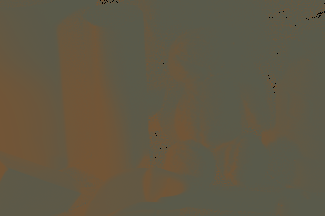}\\
\includegraphics[width=0.33\columnwidth]{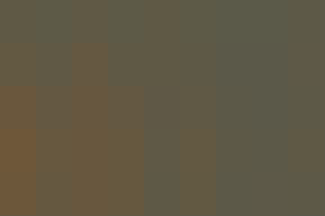}
\includegraphics[width=0.33\columnwidth]{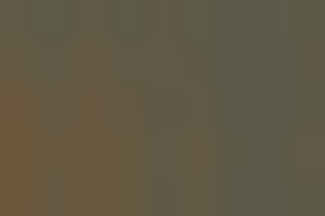}\\
\caption{Input data and results of MEM CNN for three scenes from the MIMO dataset. From top to bottom, left to right: image restored with ground-truth, image restored with estimate map, input image, ground-truth image, estimate map, filtered estimate map. The scenes images have the gamma correction applied for visualisation purposes.} 
\label{fig:outputs_FC4MI}
\end{figure}

Figure~\ref{fig:outputs_FC4MI} shows for three different scenes from the dataset, the estimate maps and the MEM CNN outputs (restored using the estimate maps), the input scenes images and the corresponding ground-truth images, and the images restored using the ground-truth. 

Some of the ground-truth images show clipped pixels. These are missed values in the calculation of the provided pixel-wise ground-truth illuminants. We do not have this clipping in the estimated results. The method output gives a good estimate of the illumination distribution in the scene. The changes in the illuminant are visible in the estimate map in the three scenes in the Figure, and these changes are accurately located in the image, even though a high frequency change in the illumination is more difficult to estimate and our solution can be improved in this direction. MEM CNN gives good visual results, compared to the restored images when the ground-truth is used.

\begin{table}
    \label{tab:tab}
    \begin{center}
      \includegraphics[width=1\columnwidth]{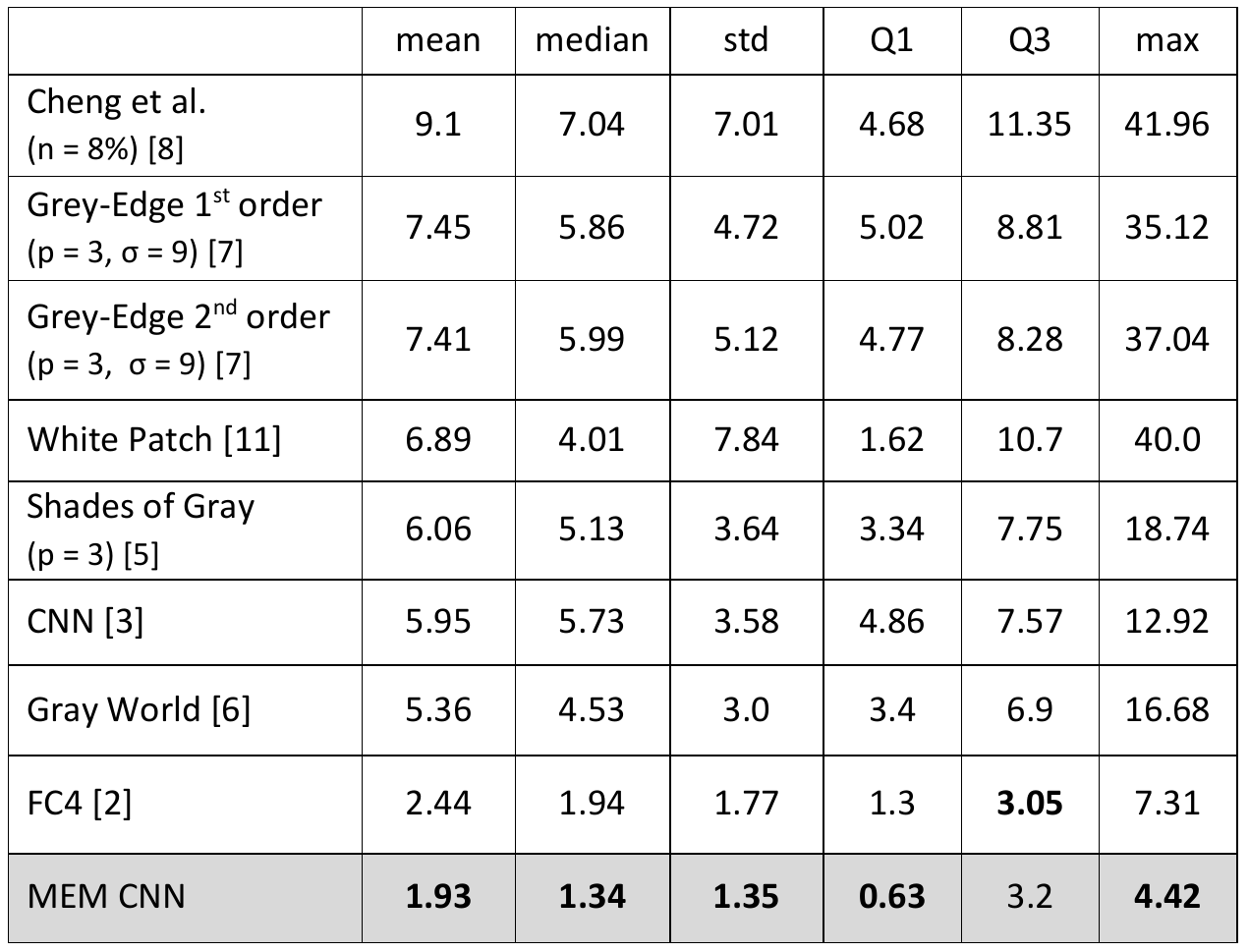}
      \caption{Table 1: Summary statistics of angular errors for different colour constancy methods evaluated on the MIMO dataset.}
    \end{center}
\end{table}

Table $1$ shows quantitative results of different methods evaluated on the MIMO dataset. They are summary statistics of the angular errors (mean, median, standard deviation, first quantile Q1, third quantile Q3 and maximum) for the whole dataset. In this multi-illuminant case, the results show that MEM CNN outperforms the other learning-based methods like CNN \cite{CNNbianco} and the original single estimate FC4 method \cite{FC4}, as well as state of the art statistics-based methods: Gray World \cite{grayworld}, White Patch \cite{whitepatch}, Shades of Gray \cite{shadesofgray}, Grey-Edges \cite{grayedge} and Cheng et al.'s method \cite{cheng2014}. 

\begin{figure}
\centering    
\includegraphics[width=0.33\columnwidth]{9_wb_gt.png}\\
\includegraphics[width=0.33\columnwidth]{9_wb_MIfc4.png}
\includegraphics[width=0.33\columnwidth]{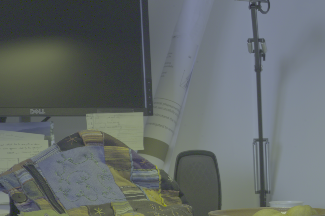}\\ 
\includegraphics[width=0.33\columnwidth]{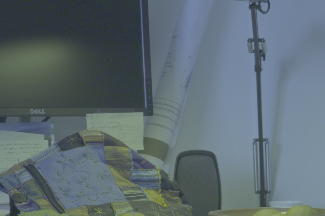}
\includegraphics[width=0.33\columnwidth]{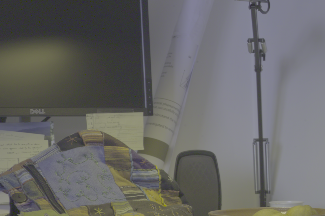}\\
\vspace{0.25cm}
\includegraphics[width=0.33\columnwidth]{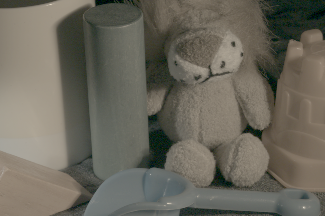}\\
\includegraphics[width=0.33\columnwidth]{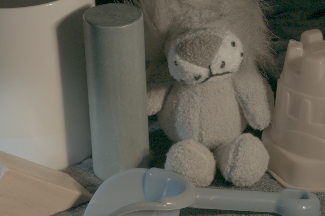}
\includegraphics[width=0.33\columnwidth]{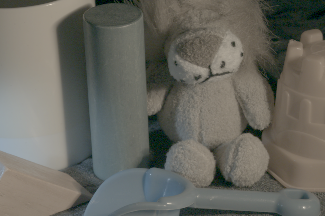}\\
\includegraphics[width=0.33\columnwidth]{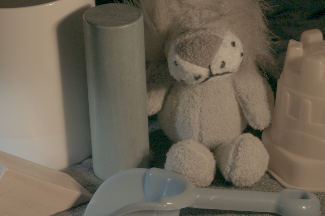}
\includegraphics[width=0.33\columnwidth]{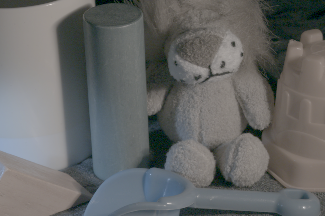}\\
\vspace{0.25cm}
\includegraphics[width=0.33\columnwidth]{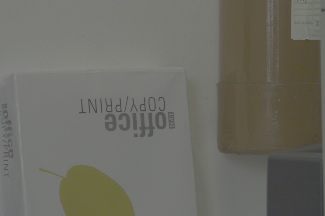}\\
\includegraphics[width=0.33\columnwidth]{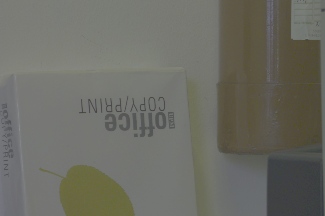}
\includegraphics[width=0.33\columnwidth]{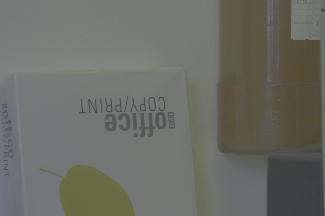}\\
\includegraphics[width=0.33\columnwidth]{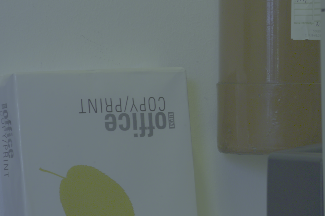}
\includegraphics[width=0.33\columnwidth]{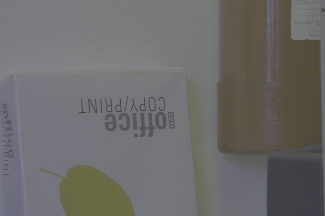}\\
\caption{Visual results of different colour constancy methods for three scenes from the MIMO dataset. From top to bottom, left to right: images restored respectively with ground-truth, MEM CNN, FC4 \cite{FC4}, CNN \cite{CNNbianco}, Gray World \cite{grayworld}. The images have the gamma correction applied for visualisation purposes.} 
\label{fig:outputs_comparison}
\end{figure}

In Figure~\ref{fig:outputs_comparison} we show outputs of three scenes using four illuminant estimation methods: MEM CNN, FC4 \cite{FC4}, CNN \cite{CNNbianco} and Gray World \cite{grayworld}, and the restored images using the ground-truth illuminants (the ground-truth image). MEM CNN gives good results. With the multi-illuminant estimation, in the given examples, our method is capable of recovering the colours of the scene under a white light without a large shift in hue, which is not the case of e.g. Gray World (first scene), FC4 (second scene) and CNN (third scene).  


\pagebreak
\section{Conclusion}
We presented in this paper an effective framework for colour constancy, MEM CNN. It is based on a deep convolutional neural network and it solves for illuminant estimation when there are multiple lights in the scene which is very often the case. MEM CNN generates a region-wise estimate map which is divided out from the image to perform colour constancy. It gives competitive results compared to state of the art and recent learning-based methods when evaluated on the MIMO multi-illuminant dataset.  

We are limited by the size of the datasets for training specifically when ground-truth illuminants images are needed. A data generator tool like in \cite{crop} can be used to create larger datasets. As a future work we would like to evaluate MEM CNN on images with more than two illuminants, where we expect it to perform well (in this work the two illuminants are not uniformly distributed in the scenes) and we would like to train the CNN on a larger dataset and improve the solution to be able to see the small and high frequency changes in the scene illumination. 






\FloatBarrier



\end{document}